# Pedestrian Trajectory Prediction with Convolutional Neural Networks

Simone Zamboni[a,1], Zekarias Tilahun Kefato[a], Sarunas Girdzijauskas[a], Norén Christoffer[b], Laura Dal Col[b]

[a]*KTH Division of Software and Computer Systems, Stockholm, Sweden*
[b]*Autonomous Transport Systems Scania CV AB, Sodertalje, Sweden*

**Abstract**

Predicting the future trajectories of pedestrians is a challenging problem that has a range of application, from crowd surveillance to autonomous driving. In literature, methods to approach pedestrian trajectory prediction have evolved, transitioning from physics-based models to data-driven models based on recurrent neural networks. In this work, we propose a new approach to pedestrian trajectory prediction, with the introduction of a novel 2D convolutional model. This new model outperforms recurrent models, and it achieves state-of-the-art results on the ETH and TrajNet datasets. We also present an effective system to represent pedestrian positions and powerful data augmentation techniques, such as the addition of Gaussian noise and the use of random rotations, which can be applied to any model. As an additional exploratory analysis, we present experimental results on the inclusion of occupancy methods to model social information, which empirically show that these methods are ineffective in capturing social interaction.

[1]Corresponding author, email: zamboni.simone@outlook.com

## 1. Introduction

Pedestrian trajectory prediction is a challenging task that is gaining increasing attention in recent years because its applications are becoming more and more relevant. These applications include human surveillance, socio-robot navigation and autonomous driving. Because these areas have become more important and demanding over time, methods to approach the problem of pedestrian trajectory prediction have evolved, transitioning from physics-based models to data-driven models that use deep learning. One of the main sources of information that these models use is the past trajectory, and thus its representation is has a great impact. Moreover, the deep learning architectures used are sequence-to-sequence, which have evolved beyond recurrent models during the last years. One of the first approaches in pedestrian behaviour modelling was introduced by Helbing et al. and it is called Social Forces Model [1]. Physics-based models like this have been extensively developed in the past, with the introduction of other techniques such as BRVO [2]. However, in recent years the data-driven approach to pedestrian behaviour modelling has become increasingly popular, thanks to its promising results. One of the most influential neural networks architecture in pedestrian trajectory prediction was introduced by Alahi et al. under the name of Social LSTM [3]. Since then several different deep learning architectures have been proposed. Common elements in these recent works are the use of Generative Adversarial Networks [4], the use of Graph Neural Networks [5], the integration of attention [6] and the inclusion of spatial [7] and image information [8].

Despite the vast number of different neural network-based approaches, there are still some unexplored aspects. The first one is data pre-processing. Pedestrian trajectory prediction models get past positions as input, however, there is no detailed study investigating if these coordinates should be normalized and what is the best normalization technique. Moreover, the total amount of publicly available data is limited, while it is widely understood that neural networks perform better with a vast amount of data. To address the issue of limited data, a solution could be to use data augmentation. However, this approach is often not explored in detail in publications. Consequently, it is currently unknown what normalization and data augmentation techniques are most effective in pedestrian trajectory prediction.

Another topic hardly explored in literature, [9] being the exception, is the use of Convolutional Neural Networks (CNN) in pedestrian trajectory prediction. In the machine translation and image caption fields it was proved, in works such as [10] and [11], that CNNs are a valid alternative to Recurrent Neural Networks (RNN). However, in pedestrian trajectory prediction, a detailed confrontation is still missing.

Consequently, the objective of this work is to find effective pre-processing techniques and to develop a convolutional model capable of outperforming models based on RNN. Models presented in this work are designed to be employed in scenarios in which only the past positions (in meters) of each pedestrian in a certain area are known. It is assumed that no information is available about the environment in which pedestrians move.

Fulfilling the outlined objectives the main contributions of this work are the following:

- The identification of effective position normalization techniques and data augmentation techniques, such as random rotations and the addition of Gaussian noise;
- The introduction of a novel model based on 2D convolutions capable of achieving state-of-the-art results on the ETH and Trajnet dataset.

In addition, we also present experimental results obtained including social information in the convolutional model. These experiments empirically show that occupancy methods are ineffective to represent social information.

The remainder of this paper is organized as follows. Section 2 introduces the related work. Section 3 introduces the problem and then presents the main techniques used, divided in data pre-processing, data augmentation, the pro-posed convolutional architecture, the recurrent baselines and techniques to add social information. Section 4 presents the datasets, the metrics, the implantation details and the results for each one of the proposed techniques in the previous section, and it is concluded with a comparison with literature results on the chosen datasets. Section 5 reports a qualitative analysis of the findings. Finally, Section 6 concludes.

## 2. Related Work

Early work from Helbing and Molnar[1] pioneered the use of physics-based models for predicting human trajectory. Their approach, the Social Forces model, considers every pedestrian as a particle subject to forces from nearby people and obstacles, and the sum of these forces gives the next pedestrian position. Physics-based pedestrian behaviour modelling has evolved over time, with the introduction of advanced techniques such as [12], [13] and BRVO [2]. These physics-based models, however, are limited by the fact that they use hand-crafted function, and thus they can represent only a subset of all possible behaviours. Deep learning models are data-driven and thus do not have this limitation. In literature, deep learning models for pedestrian trajectory prediction rely mainly on the use of Recurrent Neural Networks (RNN). One of the first works using such approach that pioneered the use of deep learning in pedestrian trajectory prediction is the Social LSTM model [3]. In this model, pedestrian trajectory together with social information is fed to an LSTM. Social information is used to model social interaction and it is represented as a grid containing nearby pedestrians.
Later works continued to use social interaction, such as in [14], but have also employed more advanced techniques, such as attention. Attention was first applied in the machine translation field [15], and one of the first work to use it for pedestrian trajectory prediction was introduced by Fernando et al. [6]. Since then multiple works have used attention in different parts of the architecture [16] [17]. A common characteristics of models based on RNN, both with and without attention, is that they are significantly more computationally expensive than other approaches, such as physics-based models and convolutional models, because of their recurrent nature.

Generative Adversarial Networks (GAN) [18] are a way to generate new synthetic data similar to training data. GAN have been seen as a way to address the multi-modal aspect of pedestrian trajectory prediction. One of the first works to use a GAN for creating multiple pedestrian trajectories was the Social GAN[4] model. In recent years the generative approach for pedestrian trajectory prediction has been extensively explored by other works using not only GAN [8] [19] [20], but also using Conditional Variational Auto-Encoders (CVAE) [21] [22]. Since generative models do not have a unique output trajectory given an input trajectory, in literature they are usually evaluated using the best-of-N method, in which N samples trajectories are generated for each input trajectory, and the prediction error is equal to the lowest error among the generated paths.

Another possible method to tackle the pedestrian trajectory prediction problem is by applying to it Graph Neural Networks (GNN). With this approach a GGN is used to describe pedestrians and their interactions through a graph: pedestrian are represented as the graph nodes while their interaction are the graph edges. One of the first works to apply GNN to pedestrian trajectory prediction was [5], followed by others like [23]. Recently, GNN have also been used to model not only social interactions but also spatial interactions, as done in works such as [24] [25] [26].

Some authors have also tried to use other available sources of information to predict the future trajectory. Some works use spatial information represented as points of interest [27] [28], as an occupancy map [7], or as a semantic segmentation of the scene [29] [30]. Meanwhile, other works use image information extracted directly from the dataset videos [8] [21] [19]. The biggest limitation for these models in undoubtedly the fact that spatial or image information is often not available, since having that type of data usually requires additional infrastructure or prior knowledge of the environment.

While significant effort has been spent on more complex modeling, in the pedes-trian trajectory prediction literature there has not been an extensive exploration of convolutional models and of data pre-processing techniques, such as data normalization and data augmentation. Therefore, this work aims to expand on the current literature by presenting effective pre-processing techniques and by proposing a novel convolutional architecture capable of outperforming more complex models.

## 3. Method

In this section, the problem is first formally presented. Then we describe different approaches to data-preprocessing, such as data normalization and data augmentation. Afterwards, the proposed convolutional architecture is presented, followed by the introduction of recurrent baselines. Finally, the chosen approaches to include social information are introduced.

### 3.1. Problem formulation

The goal of pedestrian trajectory prediction is to predict pedestrians future positions given their previous positions. Concretely, given a scene where pedestrians are present, their coordinates are observed for a certain amount of time, called $T_{obs}$, and the task is to predict the future coordinates of each pedestrian from $T_{obs}$ to $T_{pred-1}$ (assuming that time start at 0). A discretization of time is assumed, in which the time difference between time $t$ and time $t + 1$ is the same as the time difference between time $t + 1$ and time $t + 2$. The position of each pedestrian is characterized by its $(x, y)$ coordinates (in meters) with respect to a fixed point, arbitrarily selected and unique for each scene. Therefore, for pedestrian $i$ the positions $(x_t^i, y_t^i)$ for $t \in 0, ..., T_{obs-1}$ are observed and positions $(\hat{x}_t^i, \hat{y}_t^i)$ for $t \in T_{obs}, ..., T_{pred-1}$ are predicted. We denote all the past positions of a pedestrian $i$ with $X^i$, the predicted future positions with $\hat{Y}^i$ and the real future positions of pedestrian $i$ with $Y^i$. In essence, the problem of pedestrian trajectory prediction can be stated as:

*How to predict the future positions of pedestrians from their past trajectory with the lowest possible error?*

### 3.2. Data pre-processing

To effectively train a model and achieve low error rate, it is important to pre-process the data. The way this has been done is by normalizing the input coordinates and applying data augmentation techniques.

### 3.2.1. Data normalization

The input and target data of models in pedestrian trajectory prediction are coordinates, however, the origin point of these coordinates is not specified. Therefore, one might ask: *which coordinate system to use, as a form of data normalization?* To answer this question, we have identified four data-preprocessing techniques:

1. Absolute coordinates. With absolute coordinates, we refer to the naive approach: taking directly the coordinates from the datasets as they are. This is not a sensible approach since each scene has the origin point in a different position, and thus coordinates can lie in very distant intervals.

2. Coordinates with the origin in the first observation point (in essence, we impose that: $(x^i_0, y^i_0) = (0, 0)$ ). To achieve this, from each point in the sequence the first position, $(x^i_0, y^i_0)$, is subtracted. In this way, the coordinates became scene-independent and do not have the same drawbacks as absolute coordinates.

3. Coordinates with the origin in the last observation point (in essence, we impose that: $(x^i_{T_{obs-1}}, y^i_{T_{obs-1}}) = (0, 0)$ ). Similar to the previous coordinates type, but with the difference that the subtracted position is $(x^i_{T_{obs-1}}, y^i_{T_{obs-1}})$, which is the last position the network will observe.

4. Relative coordinates (velocities). In this case instead of coordinates with a fixed reference system, the network is fed with relative displacements. It is to note that if relative displacements are scaled accordingly to the annotations per seconds, they represent the instantaneous velocities.

An example of the same trajectory represented in different coordinate systems can be found in Figure 1.

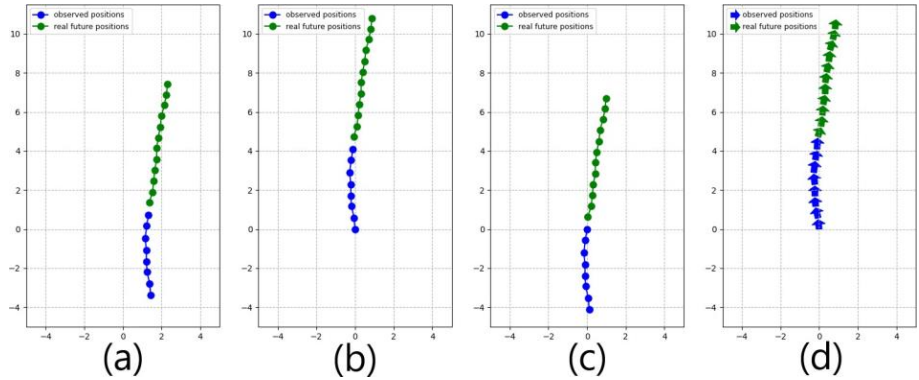

Figure 1: The same trajectory of a pedestrian going upward in the four different coordinate systems. The observed positions are in blue and the future positions are in green. (a) Absolute coordinates. (b) Origin at time $t = 0$ (the first observation point) (c) Origin at time $t = T_{obs-1}$ (the last observation point) (d) Visualization of relative coordinates(velocities) using arrows, each blue arrow is an input to the network at that timestep.

### 3.2.2. Data Augmentation

The following data augmentation techniques have been analyzed:

1. Apply a random rotation to each trajectory. This technique should make the network learn patterns in a rotation-invariant manner.
2. Mirror the trajectory on the x-axis or y-axis with a probability. No rotation applies a mirroring, therefore mirroring could enhance the effects or random rotations.
3. Apply Gaussian noise with mean 0 and standard deviation $\sigma$ to every point. Thus, at each time step the input coordinates are $(x_t^i + a, y_t^i + b)$, with $a$ and $b$ sampled at every time step from a normal distribution with mean 0 and standard deviation $\sigma$. This approach should make the network more robust to small perturbations and imprecisions.

An example of the three data augmentation techniques proposed can be found in Figure 2.

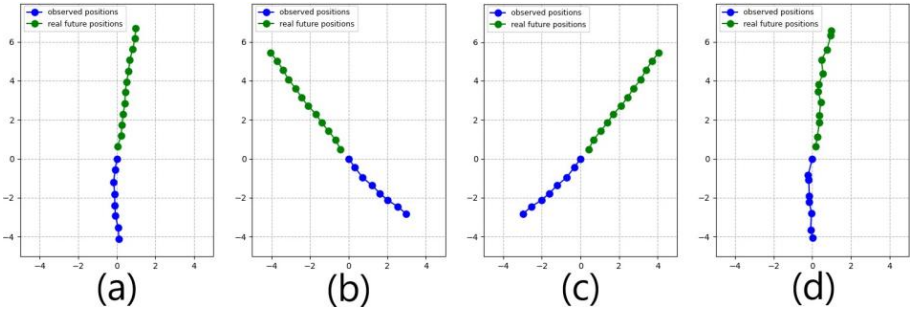

Figure 2: The effects of rotation, mirroring and noise on a trajectory. The observed positions are in blue and the future positions are in green. (a) A pedestrian trajectory with the origin in the last observation point. (b) The same pedestrian trajectory rotated (of 45 degrees). (c) The same pedestrian trajectory rotated (of 45 degrees) an mirrored (on the x-axis) (d) The pedestrian trajectory with noise of $\sigma = 0.1$ applied to each point.

### 3.3. Convolutional Model

As shown by works such as [10] and [11], CNN can be applied to problems involving sequences, such as machine translation or image captioning, achieving competitive results in comparison with RNN. It has also been shown by Nikhil and Morris in [9], that indeed a convolutional model can be employed in pedestrian trajectory prediction. However, in their architecture it is not explained in detail how to go from 8 input positions to 12 output positions, and how to transform output features in future positions. Moreover, their model does not outperform recurrent models such as SoPhie [8].

For the reasons just stated we introduce a new convolutional architecture specifically designed for pedestrian trajectory prediction. In the next paragraph the general structure of the architecture is presented, and afterwards specific models implementing this architecture are presented, together with a detailed visualization of the best one.

The proposed architecture takes 8 input positions ($(x_t^i, y_t^i)$ for $t \in 0, ..., T_{obs-1}$ and for pedestrian $i$) and outputs the future 12 positions($(\hat{x}_t^i, \hat{y}_t^i)$ for $t \in T_{obs}, ..., T_{pred-1}$ and for pedestrian $i$), as it is commonly done in the pedestrian trajectory prediction literature. As a first step each input positions is embedded in 64-length feature vector by a fully connected layer. After this first step, the input trajectory is represented by features vectors that are arranged in a 64x8 matrix, in which 64 is the embedding dimension and 8 is the number of input positions. This matrix can be interpreted either as a 64 one-dimensional channels with 8 features each, or as a one channel 64x8 image. Thus, it is possible to apply both 1D and 2D convolutions to this matrix. After the embedding, a first group of convolutions with padding is applied. The padding depends on the kernel size of the convolutions and it is employed to keep the number of features in output the same as the number of features in input. This means that as many convolutional layers as wanted can be stacked at this step. The mismatch between the input positions, which are 8, and the output positions, which are 12, require the introduction of specific layers to address this problem. Therefore, first an upsampling layer is applied to double the number of features from 8 to 16, and afterwards convolutional layers without padding are applied to reduce the number of features from 16 to 12. Lastly, a second group of convolutions with padding is applied and then a final fully connected layer transforms each feature vector in an output position.

The presented convolutional architecture is scalable, in a sense that there is no limit at the number of layers in the initial and final convolutions groups. It is also one-shot: in one pass all the output coordinates are generated, differently from recurrent models where usually one pass gives only the next position.

Multiple implementations of this generic architecture are possible. The ones explored in this work are:

1. 1D convolutional model. This is the most basic convolutional model and it interprets the 64x8 matrix created after the embedding layer as 64 one-dimensional channels with 8 features each.
2. Positional embeddings model. As proposed by [10], to give to the network the clue of order in the input data, the positional information of each input position is used.
3. Transpose convolution model, which uses transpose convolutional layers instead of the upsampling layer followed by convolutions without padding, to transition from 8 features to 12 features.

4. Residual connections model. As explored in [31], residual connections help information and gradient flow, especially in very deep architectures. In this architecture variation, all convolutional layers are transformed in residual convolutional layers.

5. 2D convolutional model. This model interprets the 64x8 matrix created after the embedding layer as one channels 64x8 image. It is important to note that 2D convolutions usually increase the number of channels, thus, the final convolutional layer needs to decrease the channels number to one so that the final fully connected layer that computes the future positions can be applied. 2D convolutions have the advantage that they process multiple features over multiple timesteps, while 1D convolutions process only one feature over multiple timesteps.

As it is possible to see in Section 4, the 2D convolutional model is the model that achieves the best results over multiple datasets, and thus it represents the main contribution of this work from an architectural point of view. The detailed architecture of the 2D convolutional model can be found in Figure 3. More information on training and hyperparameters for the all the convolutional models can be found on Section 4.3.

*3.4. Recurrent baselines*

To confront the results obtained using the convolutional model two RNN baselines have been implemented. The first is a simple LSTM. This model embeds with a fully connected layer one position ($x_t^i$, $y_t^i$) into a 64-length feature vector, which is fed to the LSTM cell and then two fully connected layers transform the output of the cell into the next position ($x_{t+1}^i$, $y_{t+1}^i$). The positions fed with t $\in$ 0, ..., $T_{obs-1}$ are the ground truth ones, while afterwards are the ones predicted by the network in the previous time step. The second baseline is an Encoder-Decoder (shorten to Enc-Dec in the tables) that uses LSTM cells both in the encoder and in the decoder. The encoder has the same architecture as the LSTM baseline except for the two fully connected layers in output which are missing, while the decoder has the exact same architecture as the LSTM baseline.

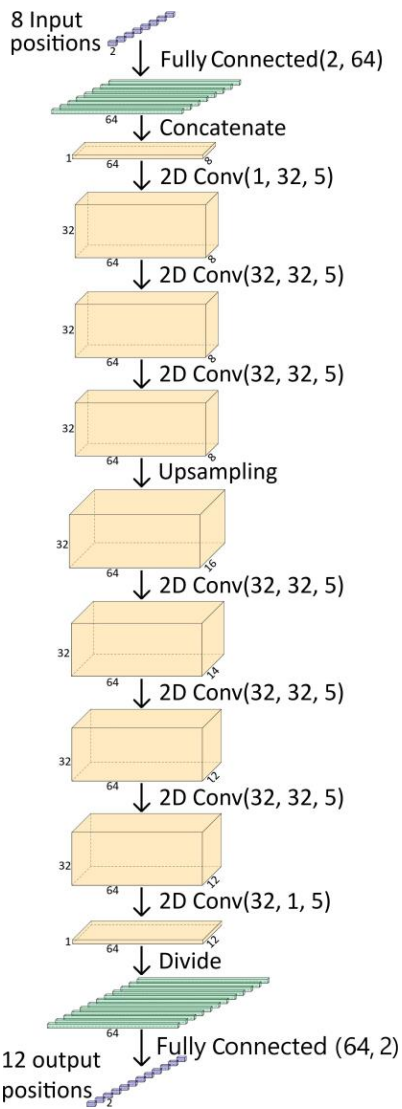

Figure 3: Convolutional 2D model. For the fully connected layers in the parenthesis there are input and output dimensions. For convolutional layers in the parenthesis there are input channels, output channels and kernel size. All the convolutions have padding 2 so that the output dimension is the same as the input dimension. The exception are the two convolutions after the upsampling layer that have a padding of 1. Each layer has a corresponding batch normalization layer. The number of layers and the kernel size was determined empirically, over multiple experiments on the two datasets presented in Section 4.

More information on the exact architecture, training and hyperparameters for the recurrent baselines can be found on Section 4.3.

*3.5. Addition of social information*

In addition to past trajectory, social information can be used as input to the network. We analyzed three simple ways to represent social information, which use the occupancies of nearby pedestrians in the space. These techniques are:

1. A square occupancy grid, introduced in Social LSTM[3].
2. A circular occupancy map, introduced in SS-LSTM[32].
3. An angular pedestrian grid, introduced in [7]. In this technique the angular space around a pedestrian is divided in a number of intervals and then the closest pedestrian in each direction, within a certain range, is computed.

A visual example of these techniques can be seen in Figure 4.

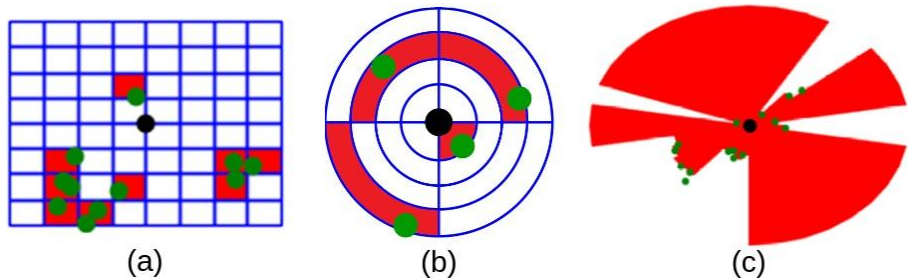

(a) (b) (c)

Figure 4: Analyzed methods to represent the occupancies of nearby pedestrians in the space. (a) Square occupancy grid. (b) Circular occupancy map. (c) Angular pedestrian grid. The current pedestrian is in black and the other pedestrians are in green. For (a) and (b) the occupied space is in red, while for (c) the free space is in red.

The square occupancy grid is represented with a matrix $l * l$ where $l$ is the number of cells on each side. The circular occupancy map is represented with a matrix $c * 4$ where $c$ is the number of circles. The angular pedestrian grid is represented by a vector of length $int(360/d)$, where $d$ is the number of degrees an element of the vector represents. Social information which is not already in vector form is flattened to be used as an input to the models.

Social information is integrated into the convolutional model and into the Encoder-Decoder baseline. Both models require minimal modifications: at each time step the social information is embedded by another fully connected layer, and then obtained social feature vector is summed to the position feature vector. This new vector represents position and social information for that timestep and it is then fed to the rest of the network. It is important to note that social information is available only during observation (therefore in the Encoder-Decoder baseline the encoder process both social and position information, while the decoder only processes position information).

## 4. Results

In this section we first describe the used datasets along with the evaluation metrics and implementation details. Then, we present the experimental results obtained training the proposed architecture and the baselines with the different data pre-processing techniques previously presented. Finally, a comparison with literature results on the two chosen datasets is displayed.

*4.1. Data*

The ETH[33] and UCY datasets[34] are two publicly available datasets widely used in literature. Jointly they contain five scenes, two from ETH (named ETH and Hotel), and three from UCY (named Univ, Zara1 and Zara2). In total, they contain more than 1600 pedestrian trajectories, with pedestrian positions annotated every 0.4 seconds. The train and test are done with the leave-one-out-cross-validation approach: a model is trained on four scenes and tested on the fifth, and this procedure is repeated five times, one for each scene. Since these two datasets are mainly used jointly from now onward the two datasets together will be referred to as the ETH-UCY dataset. The raw pedestrian positions were downloaded from the Social GAN repository [35] (which was using them to compute relative coordinates), except for the ETH scene for which the original dataset was used[33].

A more recent dataset is the Trajectory Forecasting Benchmark (also known as TrajNet) [36]. It is a curated collection of datasets, comprising in total of more than 8000 pedestrian trajectories. It merges the ETH, UCY, Stanford Drone Dataset [37] and PETS2009 [38] datasets. The Stanford Drone Dataset contributes to the majority of the pedestrian tracks. One frame is annotated with pedestrian positions every 0.4 seconds. The data has already been split in training and test by the authors, and for the test set only the observed position are available. The test error can be computed only by submitting the obtained predictions to the official dataset site [39], where a leaderboard is also present. A scene from the UCY dataset and one from Stanford Drone Dataset can be viewed in Figure 5.

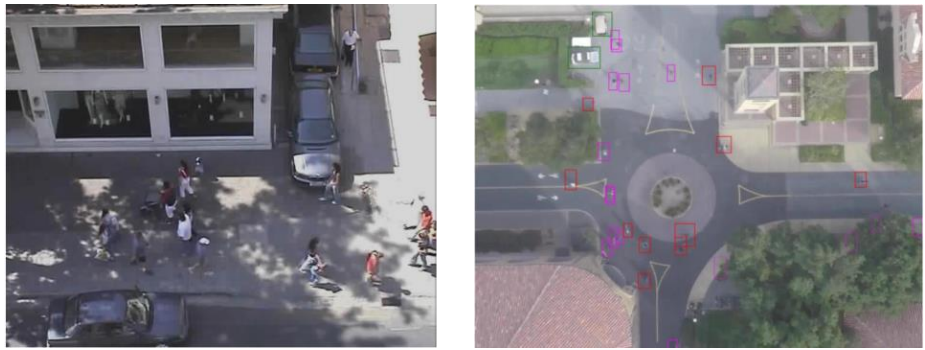

Figure 5: On the left a frame of the Zara1 scene from the UCY[34] dataset, and on the right a frame of the death_circle scene from the Stanford Drone Dataset[37].

4.2. Metrics

It is common practice in literature to set $T_{obs}$ = 8 and $T_{pred}$ = 12. Work that do this include [3] [4] [8] [9] and many others. Thus, for the sake of comparing with other models, the same setting is used in all the experiments.

The evaluation of predicted trajectories is done using metrics. The first (and most important) metric used is the Average Displacement Error (ADE), which was introduced in [33]. The ADE is the Euclidean distance over all the predicted points and the ground truth points from $T_{obs}$ to $T_{pred-1}$ averaged over all pedestrians. The ADE formula is the following:

$$ADE = \frac{\sum_{i=1}^{n} \sum_{t=T_{obs}}^{T_{pred-1}} \|\hat{Y}_t^i - Y_t^i\|}{n(T_{pred} - T_{obs})} \tag{1}$$

The number of pedestrians is $n$, the predicted coordinates for pedestrian $i$ at time $t$ are $\hat{Y}_t^i$, the real future positions are $Y_t^i$ and $\|\|$ is the Euclidean distance. The second metric used is the Final Displacement Error (FDE), which was also introduced in [33]. The FDE is the Euclidean distance between the predicted position and the real position at $t = T_{pred-1}$ averaged over all pedestrians. The FDE formula is the following:

$$FDE = \frac{\sum_{i=1}^{n} \|\hat{Y}_{T_{pred-1}}^i - Y_{T_{pred-1}}^i\|}{n} \tag{2}$$

*4.3. Implementation details*

For the ETH-UCY dataset, each network was trained for 60 epochs, with a learning rate of 0.005 and a step scheduler with gamma 0.5 and step 17. For the TrajNet dataset, each network was trained for 250 epochs, with a learning rate of 0.005 and a step scheduler with gamma 0.75 and step 35. The optimizer used was Adam. The loss used was the ADE. For the baselines, the LSTM cell size was 128, and the output dimension of the two fully connected layers in output was 64 and 2 respectively. The basic 1D convolutional model has the same number of layers as the 2D model in Figure 3. The differences lie in the number of channels, which is 64 for each layer, and the absence of batch normalization. For the Gaussian noise, the standard deviation is set to 0.05 and the mean to 0. For the mirroring, there is a 25% probability of mirroring a sample on one axis and a 50% probability of not applying any mirroring at all. For the social occupancy information, grid results are obtained using 10 cells per side ($l = 10$) and each cell with a side of 0.5m. Occupancy circle results are obtained using 12 circles ($c = 12$) 0.5m apart from each other. Angular pedestrian grid results are obtained using 8 degrees per element ($d = 8$).

*4.4. Results data pre-processing*

To show that results regarding data pre-processing are valid for both convolutional and recurrent models, the LSTM baseline and a simple 1D convolutional model (with kernel size 3) have been trained with different data pre-processing techniques.

Results obtained training the two models with different coordinate normalization approaches can be found in Table 1. The best coordinate normalization is the one in which the origin is in the last observation point, since it achieves the lowest ADE across all five scenes on both the LSTM baseline and the 1D convolutional model. This is because the last observation point is the most important one, since it is the most recent. Therefore, if the origin is placed in that position all the trajectory is seen through the lens of the most important point, and thus network better understands the whole trajectory.

|             | ETH            | Hotel           | Univ           | Zara1          | Zara2          | Average        |
|-------------|----------------|-----------------|----------------|----------------|----------------|----------------|
| Conv1D-abs  | 1.165 / 1.910  | 10.693 / 11.323 | 0.727 / 1.489  | 0.443 / 0.920  | 0.389 / 0.797  | 2.684 / 3.288  |
| Conv1D-t0   | 0.731 / 1.393  | **0.513 / 1.006** | 0.704 / 1.405 | 0.432 / 0.923  | 0.330 / 0.695  | 0.542 / **1.084** |
| Conv1D-tobs | **0.694 / 1.381** | 0.568 / 1.241 | **0.667 / 1.371** | 0.411 / 0.893 | **0.324 / 0.694** | **0.533** / 1.116 |
| Conv1D-rel  | 0.791 / 1.492  | 0.533 / 1.107   | 0.699 / 1.386  | **0.403 / 0.864** | 0.327 / 0.696 | 0.550 / 1.109  |
| LSTM-abs    | 7.499 / 7.961  | 6.769 / 8.733   | 1.278 / 2.061  | 0.498 / 1.042  | 0.464 / 0.962  | 3.302 / 4.152  |
| LSTM-t0     | 0.779 / 1.509  | 0.546 / 1.101   | 0.729 / 1.452  | 0.425 / **0.915** | **0.318 / 0.702** | 0.559 / 1.136 |
| LSTM-tobs   | **0.734 / 1.432** | **0.501 / 1.053** | **0.687 / 1.430** | 0.424 / 0.920 | 0.330 / 0.719 | **0.535 / 1.111** |
| LSTM-rel    | 0.747 / 1.450  | 0.589 / 1.186   | 0.688 / 1.447  | 0.445 / 0.951  | 0.325 / 0.708  | 0.558 / 1.149  |

Table 1: Base 1D convolutional model and LSTM baseline trained with different coordinate normalization techniques. Regarding the naming system, 'abs' stands for absolute coordinates, 't0' stands for coordinates with the origin in the first observation point, 'tobs' for coordinates with the origin in the last observation point and 'rel' for relative coordinates. Results are in the format ADE / FDE and the best results (for each type of network) are in bold.

Regarding the data augmentation techniques, their effects are shown in Table 2. Results show that mirroring (shorten to M in Table 2) proved to be ineffective as a stand-alone technique. Gaussian noise(N in Table 2), instead, proved effective, but the lowest average error achieved by a single data augmentation technique was obtained by random rotations(R in Table 2). We also tried all the possible data augmentation techniques combinations, however the ones includ-ing mirroring, such as MN, MR, MNR, showed no improvements with respect to N, R and NR respectively (hence results from MN, MR and MNR are omitted for brevity in Table 2 ). The lowest average error is achieved by the NR (noise and rotations) variation, even if in some scenes the error actually increases if compared with only noise or only rotations. Thus, we can affirm that mirroring is ineffective as a data augmentation technique both alone and together with other techniques. But most importantly, we can conclude that adding Gaussian noise with mean 0 to every point and applying random rotations to the whole trajectory significantly lowers the average prediction error.

|  | ETH | Hotel | Univ | Zara1 | Zara2 | Average |
|---|---|---|---|---|---|---|
| Conv1D | 0.694 / 1.381 | 0.568 / 1.241 | 0.667 / 1.371 | 0.411 / **0.893** | **0.324 / 0.694** | 0.533 / 1.116 |
| Conv1D-M | 0.690 / 1.386 | 0.599 / 1.222 | 0.673 / 1.372 | **0.409** / 0.889 | 0.330 / 0.701 | 0.532 / 1.114 |
| Conv1D-N | **0.592** / 1.220 | 0.445 / 1.011 | 0.669 / 1.375 | 0.424 / 0.903 | 0.337 / 0.720 | 0.493 / 1.046 |
| Conv1D-R | 0.668 / 1.296 | 0.318 / 0.603 | **0.576 / 1.210** | 0.471 / 1.046 | 0.349 / 0.763 | 0.476 / 0.983 |
| Conv1D-NR | 0.605 / **1.190** | **0.264 / 0.509** | 0.588 / 1.241 | 0.521 / 1.095 | 0.351 / 0.755 | **0.466 / 0.958** |
| LSTM | 0.734 / 1.432 | 0.501 / 1.053 | 0.687 / 1.430 | **0.424 / 0.920** | 0.330 / 0.719 | 0.535 / 1.111 |
| LSTM-M | 0.741 / 1.440 | 0.495 / 1.041 | 0.679 / 1.421 | 0.427 / 0.925 | 0.331 / 0.721 | 0.535 / 1.110 |
| LSTM-N | 0.621 / 1.249 | 0.421 / 0.865 | 0.698 / 1.447 | 0.428 / 0.917 | 0.334 / **0.712** | 0.500 / 1.038 |
| LSTM-R | 0.689 / 1.331 | 0.305 / 0.576 | **0.549 / 1.199** | 0.439 / 0.971 | **0.329** / 0.728 | 0.462 / 0.961 |
| LSTM-NR | **0.581 / 1.168** | **0.259 / 0.503** | 0.578 / 1.241 | 0.463 / 1.022 | 0.346 / 0.748 | **0.446 / 0.936** |

Table 2: Base 1D convolutional model and LSTM baseline trained with different data augmentation techniques. Regarding the naming system, N stands for Gaussian noise, R for random rotations and M for mirroring. The coordinates used are the ones with the origin in the last observation point (tobs). Results are in the format ADE / FDE and the best results (for each type of network) are in bold.

As Table 1 and Table 2 clearly show, results on data pre-processing techniques are valid both for convolutional and recurrent models, and this demonstrates that these findings are applicable to a multitude of architectures. In fact, the same conclusions can be obtained training the Encoder-Decoder baseline and other convolutional model variations (results omitted for brevity).

It is also interesting to note that the LSTM baseline together with data augmentation outperformed the 1D convolutional model with kernel size 3, however this is not the case with other convolutional models, as Section 4.5 shows.

*4.5. Results convolutional model variations and baselines*

Results obtained with different convolutional model variations (and baselines) are shown in Table 3. These results suggest that models with a bigger kernel size are able to generate more refined predictions, since the 1D convolutional model with kernel size 7 obtains better results than the same model with kernel size 3. The intuition behind why a bigger kernel size might be better is that the more information a kernel can process the better it can interpret complex behaviours in the trajectory. This idea still applies when the 1D convolution model is confronted with the 2D convolution model. In the first, the kernel looks at the same feature on multiple timesteps. In the second, instead, the kernel looks at multiple features in multiple timesteps and thus it process more information and generates better predictions. However, this intuition has diminishing returns: experiments with the 2D convolutional model using kernel size 7 generated slightly worst results compared to the same 2D model with kernel size 5.

Regarding other convolutional model variations, using positional embedding and transpose convolutions proved to be ineffective. Moreover, adding residual connections also did not improve results, since the optimal number of convolutional layers is quite limited (7, as Figure 3 shows) and thus residual connections are not needed.

Table 3 also offers a comparison between the baselines and the proposed convolutional models. The 1D convolutional model is able and outperform the recurrent baselines only when using a bigger kernel size, while the best model is the 2D convolutional with kernel size 5. Thus, we can conclude that it is indeed possible to develop a convolutional model capable of outperforming recurrent models in pedestrian trajectory prediction. However, it is interesting to note that the difference in average error between the recurrent baselines and the convolutional models is not ample.

|  | ETH | Hotel | Univ | Zara1 | Zara2 | Average |
|---|---|---|---|---|---|---|
| Conv1D-Ks3 | 0.605 / 1.190 | 0.264 / 0.509 | 0.588 / 1.241 | 0.521 / 1.095 | 0.351 / 0.755 | 0.466 / 0.958 |
| Conv1D-Ks7 | 0.560 / 1.149 | 0.246 / 0.427 | 0.590 / 1.249 | 0.478 / 1.046 | **0.346 / 0.737** | 0.444 / 0.931 |
| Conv1D-Ks7+Pe | 0.568 / 1.125 | 0.248 / 0.467 | 0.594 / 1.257 | 0.459 / 0.990 | 0.369 / 0.789 | 0.447 / 0.926 |
| Conv1D-Ks7+Rc | 0.606 / 1.197 | 0.267 / 0.517 | 0.595 / 1.254 | **0.451 / 0.989** | 0.356 / 0.762 | 0.455 / 0.944 |
| Conv1D-Ks7+Tc | 0.560 / 1.121 | 0.245 / 0.470 | 0.589 / 1.251 | 0.516 / 1.073 | 0.349 / 0.741 | 0.452 / 0.931 |
| Conv2D-Ks5 | **0.559 / 1.114** | **0.240 / 0.464** | 0.581 / 1.225 | 0.456 / 0.993 | 0.347 / 0.751 | **0.436 / 0.909** |
| LSTM | 0.581 / 1.168 | 0.259 / 0.503 | **0.578 / 1.214** | 0.463 / 1.022 | 0.347 / 0.748 | 0.446 / 0.936 |
| EncDec | 0.585 / 1.170 | 0.246 / 0.491 | 0.589 / 1.245 | 0.467 / 1.023 | 0.360 / **0.737** | 0.449 / 0.938 |

Table 3: Convolutional models variants and baselines compared. Regarding the naming system, Ks denotes the kernel size, Pe stands for Positional embeddings, Rc for Residual connections and Tc for transpose convolutions. These networks are trained with random rotations, Gaussian noise and coordinates with the origin in the last observation point(tobs-NR). Results are in the format ADE / FDE and the best results are in bold.

*4.6. Results using social information*

Results of Table 4, in which the 2D convolutional model is trained with social information, are unexpected: the addition of social information proved to be ineffective on the ETH-UCY dataset. Similar results are also obtained with the Encoder-Decoder baseline: architectures that use the proposed social occupancy information methods are not able to outperform the same architectures without social information. This is indicated by the fact that networks with social information obtain very similar results to networks without it, as occupancy information would not be relevant. Upon further investigation, it was found that the average gradient flow in the social information embedding weights of the networks was around 50-100 times smaller than the average gradient flow in the position embedding weights. This might suggest that for the network there is very little correlation between the real future trajectory and social information, and thus this kind of information is almost ignored. An example of the gradient flow in the network can be found in Figure 6.

Results on the addition of social information are to be considered mainly as an exploratory analysis. Much more can be done (and has been done) to include social information as input to a model in pedestrian trajectory prediction. What our results show is that the specific approaches that use occupancy infor-mation that we tested, in combination with the presented architectures, failed to improve results on the ETH-UCY dataset.

|              | ETH           | Hotel         | Univ          | Zara1         | Zara2         | Average       |
|--------------|---------------|---------------|---------------|---------------|---------------|---------------|
| Conv2D + Sog | **0.558** / 1.118 | **0.233** / 0.455 | 0.604 / 1.269 | 0.464 / 1.005 | 0.342 / 0.740 | 0.440 / 0.915 |
| Conv2D + Com | 0.561 / 1.122 | 0.235 / **0.447** | 0.590 / 1.240 | 0.461 / **0.991** | 0.348 / 0.746 | 0.439 / 0.910 |
| Conv2D + Apg | 0.567 / **1.109** | 0.235 / 0.449 | 0.589 / 1.231 | 0.464 / 0.997 | **0.337 / 0.719** | 0.438 / **0.901** |
| Conv2D       | 0.559 / 1.114 | 0.240 / 0.464 | **0.581 / 1.225** | **0.456** / 0.993 | 0.347 / 0.751 | **0.436** / 0.909 |

Table 4: 2D convolutional models with social information comparison. Regarding the naming system, Sog stands for Square occupancy grid, Com for Circular occupancy map and Apg for Angular pedestrian grid. These networks are trained with random rotations, Gaussian noise and coordinates with the origin in the last observation point(tobs-NR). Results are in the format ADE / FDE and the best results are in bold.

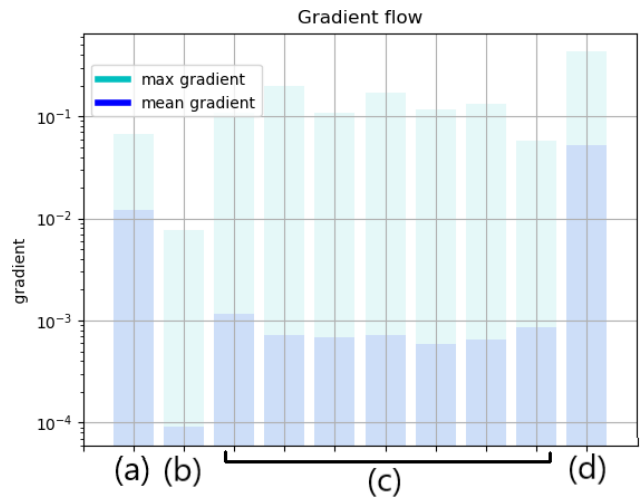

Figure 6: Example of gradient flow in the 2D convolutional model that uses a square occupancy grid to represent social information. On the x axis there are the layers of the network, and on the y axis the gradient (mean and max on a logarithmic scale) that is computed during the backpropagation after a mini-batch. (a) Position embedding layer (b) Square occupancy grid embedding layer (c) Convolutional layers (d) Final fully connected layer.

### 4.7. Comparison with literature on the ETH-UCY dataset

The following models from literature have been chosen to do a comparison with the results obtained on the ETH-UCY dataset:

- Linear Velocity, a linear regressor that estimates linear parameters by minimizing the least square error, taken from [4].

- A simple LSTM, trained by [4];

- Social-LSTM [3], trained by [4];

- Convolutional Neural Networks for trajectory prediction (shorten to CNN in the table) [9], convolutional model developed by Nikhil and Morris;

- Social-GAN [4], a generative model that uses social information with relative coordinates as data normalization;

- SoPhie [8], a generative model that uses both social and image information;

- Stochastic Trajectory Prediction with Social Graph Network (Stochastic GNN) [23], generative model that uses social information and GNN;

- MCENET [21], generative model based on a CVAE that uses both social and image information;

- Conditional Generative Neural System (CGNS) [22], generative model based on a CVAE that uses both social and image information;

- Social-BiGAT [19], generative model that uses both social and image information;

- SR-LSTM [40], model based on the state refinement of the LSTM cells of all the pedestrians in the scene to take into account for social interaction, which uses both coordinates with the origin in the last observation point and random rotations as data pre-processing;

- Social Spatio-Temporal Graph Convolutional Neural Network (STGCNN) [26], generative model that uses social information and GNN;

- STGAT [24], generative model that uses social information and GNN;

- Trajectron++ [41], a graph-structured recurrent model that also uses map information, results are from the deterministic most likely version.

|  | ETH | Hotel | Univ | Zara1 | Zara2 | Average |
|---|---|---|---|---|---|---|
| Linear Velocity, from [4] | 1.33 / 2.94 | 0.39 / 0.72 | 0.82 / 1.59 | 0.62 / 1.21 | 0.77 / 1.48 | 0.79 / 1.59 |
| Social LSTM [3],from [4]) | 1.09 / 2.35 | 0.79 / 1.76 | 0.67 / 1.40 | 0.47 / 1.00 | 0.56 / 1.17 | 0.72 / 1.54 |
| LSTM (from [4]) | 1.09 / 2.41 | 0.86 / 1.91 | 0.61 / 1.31 | 0.41 / 0.88 | 0.52 / 1.11 | 0.70 / 1.52 |
| CNN [9] | 1.04 / 2.07 | 0.59 / 1.17 | 0.57 / 1.21 | 0.43 / 0.90 | 0.34 / 0.75 | 0.59 / 1.22 |
| Social GAN∗ [4] | 0.81 / 1.52 | 0.72 / 1.61 | 0.60 / 1.29 | 0.34 / 0.69 | 0.42 / 0.84 | 0.58 / 1.18 |
| Sophie∗ [8] | 0.70 / 1.43 | 0.76 / 1.67 | 0.54 / 1.24 | **0.30** / 0.63 | 0.38 / 0.78 | 0.54 / 1.15 |
| Stochastic GNN∗ [23] | 0.75 / 1.63 | 0.63 / 1.01 | 0.48 / 1.08 | **0.30** / 0.65 | 0.26 / 0.57 | 0.49 / 1.01 |
| MCENET∗ [21] | 0.75 / 1.61 | 0.37 / 0.68 | 0.58 / 1.18 | 0.33 / 0.65 | **0.23** / 0.49 | 0.49 / 0.98 |
| CGNS∗ [22] | 0.62 / 1.40 | 0.70 / 0.93 | 0.48 / 1.22 | 0.32 / 0.59 | 0.35 / 0.71 | 0.49 / 0.97 |
| Social-BiGAT∗ [19] | 0.69 / 1.29 | 0.48 / 1.01 | 0.55 / 1.32 | **0.30** / 0.62 | 0.36 / 0.75 | 0.48 / 1.00 |
| SR-LSTM [40] | 0.63 / 1.25 | 0.37 / 0.74 | 0.51 / 1.10 | 0.41 / 0.90 | 0.32 / 0.70 | 0.45 / 0.94 |
| Social STGCNN∗[26] | 0.64 / **1.11** | 0.49 / 0.85 | **0.44 / 0.79** | 0.34 / **0.53** | 0.30 / **0.48** | 0.44 / **0.75** |
| STGAT∗ [24] | 0.65 / 1.12 | 0.35 / 1.12 | 0.52 / 1.10 | 0.34 / 0.69 | 0.29 / 0.60 | 0.43 / 0.83 |
| Trajectron++ [41] | 0.71 / 1.66 | **0.22/ 0.46** | **0.44** / 1.17 | **0.30** / 0.79 | **0.23** / 0.59 | **0.38** / 0.93 |
| LSTM-tobs-NR † | 0.581 / 1.168 | 0.259 / 0.503 | 0.578 / 1.214 | 0.463 / 1.022 | 0.346 / 0.748 | 0.446 / 0.936 |
| EncDec-tobs-NR † | 0.585 / 1.170 | 0.246 / 0.491 | 0.589 / 1.245 | 0.467 / 1.023 | 0.360 / 0.771 | 0.449 / 0.938 |
| Conv1D-tobs-NR-Ks7 † | 0.560 / 1.190 | 0.246 / 0.472 | 0.590 / 1.249 | 0.478 / 1.046 | 0.346 / 0.737 | 0.444 / 0.926 |
| Conv2D-tobs-NR-Ks5 † | **0.559 / 1.114** | 0.240 / **0.464** | 0.581 / 1.225 | 0.456 / 0.993 | 0.347 / 0.751 | 0.436 / 0.909 |

Table 5: Comparison with literature results on the ETH-UCY dataset. The results of models that have a reference have been taken directly from the publication. Generative models evaluated with the best-of-N approach with N=20 are denoted with ∗, while models developed in this work are denoted with †. Results are in the format ADE / FDE and the best results are in bold.

It is to note that since generative models have stochastic outputs, in literature they are evaluated using the best-of-N method. With this approach, N samples trajectories (for each input trajectory) are generated, and the ADE and FDE are evaluated only on the generated path with the lowest error. The value of N usually set to 20 in literature.

The result comparison for the ETH-UCY dataset can be found in Table 5. In there, the 2D convolutional model achieves the lowest error across the whole ETH dataset and an average error on the whole ETH-UCY dataset comparable to the STGAT and STGCNN models. On the UCY dataset, however, other models surpass the 2D convolutional model such as Trajectron++. This might be due to the fact that in the ETH dataset there is less pedestrian density, while in the UCY dataset there are more pedestrians per scene and thus social interaction, which is not taken into account by the 2D convolutional model, is more important. The recurrent baselines also achieve a very low error, especially if our LSTM-tobs-NR is compared to the LSTM trained by [4], thanks to the employed data pre-processing techniques.

*4.8. Comparison with literature on the TrajNet dataset*

The following models from literature have been chosen to do a comparison with the results obtained on the TrajNet dataset:

- Social LSTM [3], results are taken from the TrajNet site;
- Social GAN [4], results are taken directly from the TrajNet site;
- Location-Velocity Attention [16], model that uses location and velocity in two different LSTM with an attention layer between the two cells, the results are taken directly from the paper;
- Social Forces model [1], with results taken from the TrajNet site and from [42];
- SR-LSTM [40], the results are taken directly from the TrajNet site;
- RED (v3 from the TrajNet site), the best model from [42].

In particular, a detailed comparison with RED [42] can highlight in which ways our approach differs from previous literature and consequently how it is able to achieve a lower error. Starting from the architectural point of view, RED is a recurrent encoder with a dense multi-layer perceptron stacked on top. Our LSTM-tobs-NR has a similar architecture, since RED also uses a LSTM cell. Our convolutional model, on the other hand, has a completely different architecture since it uses convolutional layers and it is not recurrent. Regarding data normalization, RED uses relative coordinates, while our models use coordinates with the origin in the last observation point, since we empirically showed (in Table 1 ) that they produce better results. However, the biggest difference between our approach and RED is in the data augmentation. The only data augmentation in RED is the reversing of the trajectories, which doubles the amount of possible training data. However, applying random rotations and noise as we propose can transform a single trajectory in virtually infinite ways, achieving more diversity in the training data and leading to a reduced error.

The result comparison for the TrajNet dataset can be found in Table 6.

|  | ADE | FDE |
|---|---|---|
| Social LSTM [3] | 0.675 | 2.098 |
| Social GAN [4] | 0.561 | 2.107 |
| Location-Velocity Attention [16] | 0.438 | 1.449 |
| Social Forces [1] (from [42]) | 0.371 | 1.266 |
| SR-LSTM [40] | 0.370 | 1.261 |
| EncDec-tobs-NR + Sog † | 0.369 | 1.231 |
| Conv2D-tobs-NR-Ks5 + Apg † | 0.366 | 1.223 |
| Conv1D-tobs-NR-Ks7 † | 0.365 | 1.220 |
| EncDec-tobs-NR + Apg † | 0.364 | 1.218 |
| EncDec-tobs-NR † | 0.362 | 1.220 |
| Conv2D-tobs-NR-Ks5 + Sog † | 0.360 | 1.215 |
| RED (v3) [42] | 0.360 | 1.201 |
| LSTM-tobs-NR † | 0.356 | 1.212 |
| **Conv2D-tobs-NR-Ks5 †** | **0.352** | **1.192** |

Table 6: Comparison of different models on the TrajNet dataset. Models from this work are trained with random rotations, Gaussian noise and coordinates with the origin in the last observation point(tobs-NR) and denoted with †. Best model in bold.

We can affirm that the 2D convolutional model achieves state-of-the-art performances on the TrajNet dataset, making it the model with the lowest ADE on the biggest publicly available dataset for pedestrian trajectory prediction. Our LSTM-tobs-NR also achieves a very low error, lower than RED thanks to the proposed data pre-processing techniques . Finally, also on the TrajNet data the analyzed techniques for modelling social interaction proved to be ineffective (results using a circular occupancy map are missing in Table 6 because their results are very similar to the square occupancy grid). In fact, both the 2D convolutional model and the Encoder-Decoder baseline outperform their variants that use social information.

## 5. Discussion

The ADE and the FDE are not the only aspects that can be taken into consideration when evaluating a pedestrian trajectory prediction model. Other characteristics are the computational time and the number of hyperparameters. These aspects are discussed in the first part of this section.
Additionally, the accuracy of a model can depend on the situation it is trying to predict. Thus, for future improvements, it is important to understand in which scenarios the proposed architecture fails. This topic is discussed in the second part of this section.

### 5.1. Convolutional model and recurrent models comparison

Analyzing the recurrent baselines and the convolutional model beyond their quantitative results, three main differences have emerged. The first is computation time. As can be seen in Table 7, the convolutional model is more than three times faster than the Encoder-Decoder baseline and more than four times faster than the LSTM baseline at test time. These results are also valid during training time. Thus, the convolutional model is not only more accurate but also more efficient than the recurrent baselines.

|  | batch size=1 | batch size=32 |
|---|---|---|
| Convolutional 2D model | **0.0033s** | **0.00017s** |
| (155k parameters) | **per element** | **per element** |
| LSTM baseline | 0.0207s | 0.00064s |
| (106k parameters) | per element | per element |
| Encoder-Decoder baseline | 0.0118s | 0.00043s |
| (208k parameters) | per element | per element |

Table 7: Comparison of the computational test time of different models on an Nvidia Quadro 1000.

The second difference between the recurrent models and the convolutional model is the number of hyperparameters. The LSTM and Encoder-Decoder baselines have a very small number of hyperparameters (embedding size, hidden state length and the output fully connected layers dimension). Meanwhile, the convolutional model has a bigger number of hyperparameters (embedding size, number of layers, number of channels for each layer and kernel size for each layer). Therefore, the convolutional model requires more hyperparameter tuning than the recurrent models.

The third difference is flexibility. A recurrent model can be trained to observe, for example, 6 positions and predict the next 16 without any change in the architecture. It is also possible to train a recurrent model to give predictions after observing a variable number of inputs without any change in the architecture. This is not true in the case of the convolutional model. To change the number of input or output positions in the convolutional model some adjustments need to be done, mainly revolving around the upsampling layer and the convolutional layers without padding. Regarding using a convolutional model with a variable number of inputs for pedestrian trajectory prediction, that is an open challenge and might be an interesting direction for the future work.

We can therefore conclude that the convolutional model is more efficient and accurate than the recurrent baselines, but it is less flexible and requires more hyperparameter tuning.

*5.2. Failure cases*

In some of the applications of pedestrian trajectory prediction, such as autonomous driving, is important to not only to have a small average error but also to have a small maximum error. How well the proposed 2D convolutional model satisfies this constraint can be seen looking at the distribution of the Average Displacement Error in Figure 7. There, it is possible to note that the prediction error distribution resembles a Gaussian curve with a long tail.

Analyzing the poor predictions in the long tail we discovered three scenarios in which the prediction error is consistently high:

1. Sharp turns. In this case, the typical scenario is the following: a person is going straight and then does a 90-degree turn because the road was

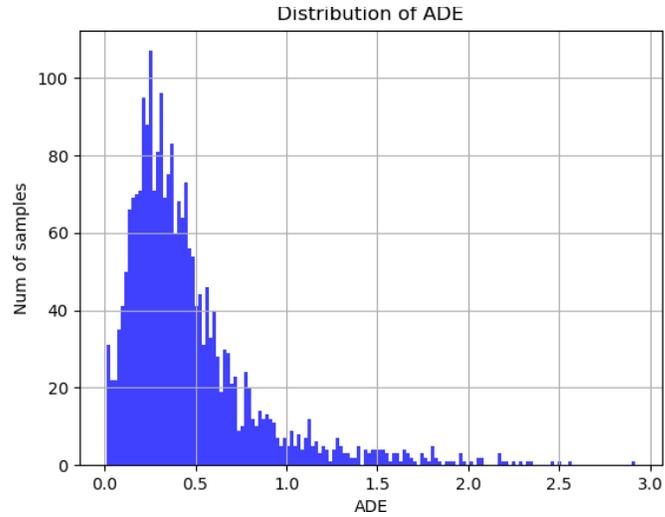

Figure 7: ADE distribution obtained testing the 2D convolutional model on the Zara1 scene. The average is 0.456, with a standard deviation of 0.370. The biggest value is 2.92.

    either turning or forking. An example of such behaviour can be seen in Figure 8. In scenarios like this, it is reasonable to assume that only models including spatial information can predict the turn reliably. What models that do not include spatial information can learn is to adapt quickly to sharp changes in trajectory, as shown in Figure 8.

2. Pedestrians stopping. In this case, it is often difficult to understand the reasons for this kind of behaviour: a person could stop to look at some shops windows, to check before crossing the street, to greet some friends, or to simply wait for someone else. Spatial information could help on some of these scenarios, but not in all.

3. Pedestrians that resume walking after stopping. This kind of behaviour happens after the previous one, and it is even more difficult to predict. If a person is still it is very difficult to understand the exact moment when it will resume moving. The safest assumption is that the pedestrian will continue to remain still, which leads to a very high error if the network observation ends a few moments before the person starts walking.

Analyzing these three scenarios it is possible to affirm that, to reduce instances in which the error is very high, the inclusion of spatial information could be very effective. Consequently, as a future work, the inclusion of spatial information in the convolutional model appears to be a promising direction.

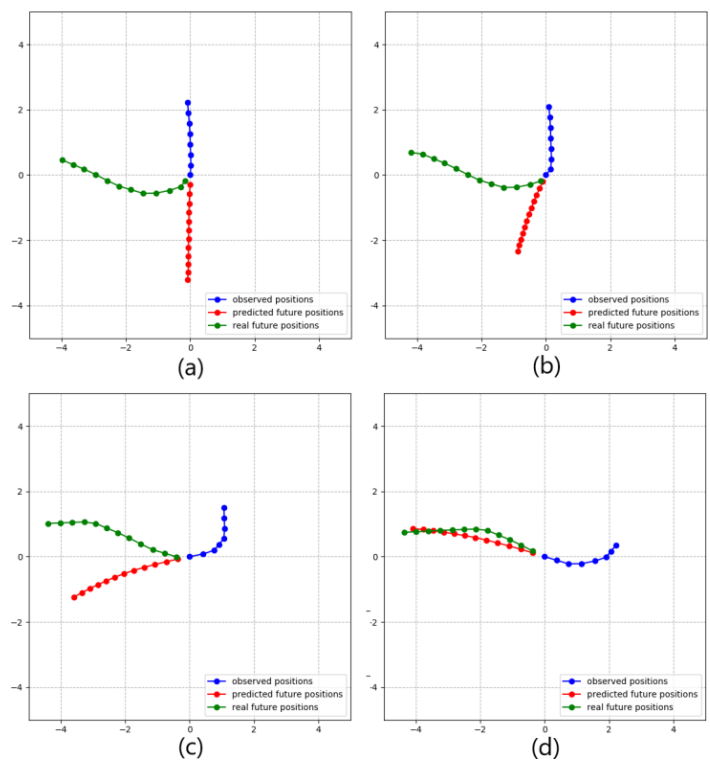

Figure 8: (a) The worst prediction of the 2D convolutional model on the Zara1 scene: a person is going down and then changes direction very sharply. The prediction is inaccurate because in the observed positions there are no clues of a turn. (b) The prediction on the same person one frame (0.4 seconds) after: with only one position pointing in a different direction the network is able to understand that the pedestrian is turning. (c) The trajectory after other three frames: the prediction aligns even more with reality. (d) After other three frames the prediction is very similar to the ground truth.

Accepted Manuscript to Pattern Recognition                                                                 29

## 6. Conclusion

In this work, we first confronted various data pre-processing techniques for pedestrian trajectory prediction. We found out that the best combination is obtained using coordinates with the origin in the last observation point as data normalization and applying Gaussian noise and random rotations as data augmentation. This solution proved to be effective in multiple architectures, both convolutional and recurrent, demonstrating that these findings are general thus can benefit future works in the pedestrian trajectory prediction field.

We also proposed a new convolutional model for pedestrian trajectory prediction that uses 2D convolution. This new model is able to outperform the recurrent baselines, both in average error and in computational time, and it achieves state-of-the-art results on the ETH and TrajNet datasets.

As an additional exploratory analysis, we also presented empirical results on the inclusion of social occupancy information. Our results suggest that the inclusion of social occupancy information does not reduce the prediction error.

Accompanying these quantitative results, a comparison between the convolutional and recurrent models was presented. Our analysis concluded that the convolutional models can be more efficient and accurate than the recurrent baselines, but are less flexible and require more the hyperparameter tuning.

Moreover, an analysis of the most common failure scenarios in the predictions has been carried out, pointing out that the most common scenarios with high prediction error are pedestrians doing sharp turns, pedestrians stopping and pedestrians that resume walking after stopping.

Interpreting these finding one can see as a promising direction for future work the inclusion of spatial information as input to the convolutional model, to address the pedestrians turning. Another interesting future direction is a better inclusion of social information using more advanced techniques, that should be specifically designed to be applied to a convolutional architecture. A relevant dataset to evaluate these findings could be TrajNet++ [43], which is specifically designed to assess social interaction and collisions.

**Acknowledgements**

This work is the result of Simone Zamboni's master thesis project carried out at SCANIA Autonomous Transport Systems. We thank the support of the industry partner, SCANIA, and the support of the university partner, KTH.

©2021. This manuscript version is made available under the CC-BY-NC-ND 4.0 license
         https://creativecommons.org/licenses/by-nc-nd/4.0/
Article DOI: https://doi.org/10.1016/j.patcog.2021.108252

[7] M. Pfeiffer, G. Paolo, H. Sommer, J. Nieto, R. Siegwart, C. Cadena, A data-driven model for interaction-aware pedestrian motion prediction in object cluttered environments, in: 2018 IEEE International Conference on Robotics and Automation (ICRA), 2018, pp. 1–8.

[8] A. Sadeghian, V. Kosaraju, A. Sadeghian, N. Hirose, H. Rezatofighi, S. Savarese, SoPhie: An attentive GAN for predicting paths compliant to social and physical constraints, in: Proceedings of the IEEE Conference on Computer Vision and Pattern Recognition (CVPR), 2019, pp. 1349–1358.

[9] N. Nikhil, B. T. Morris, Convolutional neural networks for trajectory prediction, in: Proceedings of the European Conference on Computer Vision (ECCV), 2018.

[10] J. Gehring, M. Auli, D. Grangier, D. Yarats, Y. Dauphin, Convolutional sequence to sequence learning, in: arXiv preprint, Vol. arXiv/1705.03122, 2017.

[11] J. Aneja, A. Deshpande, A. G. Schwing, Convolutional image captioning, in: Proceedings of the IEEE Conference on Computer Vision and Pattern Recognition (CVPR), 2018, pp. 5561–5570.

[12] Z. Chen, L. Wang, N. H. Yung, Adaptive human motion analysis and prediction, in: Pattern Recognition, Vol. 44, 2011, pp. 2902–2914.

[13] C. Barata, J. C. Nascimento, J. M. Lemos, J. S. Marques, Sparse motion fields for trajectory prediction, in: Pattern Recognition, Vol. 110, 2021, p. 107631.

[14] Z. Pei, X. Qi, Y. Zhang, M. Ma, Y.-H. Yang, Human trajectory prediction in crowded scene using social-affinity long short-term memory, in: Pattern Recognition, Vol. 93, 2019, pp. 273–282.

[15] D. Bahdanau, K. Cho, Y. Bengio, Neural machine translation by jointly learning to align and translate, in: ArXiv, Vol. arXiv:1409.0473, 2014.
32Accepted Manuscript to Pattern Recognition
©2021. This manuscript version is made available under the CC-BY-NC-ND 4.0 license
    https://creativecommons.org/licenses/by-nc-nd/4.0/
Article DOI: https://doi.org/10.1016/j.patcog.2021.108252